\documentclass[a4paper,conference]{IEEEtran}
\IEEEoverridecommandlockouts
\usepackage{cite}
\usepackage{amsmath,amssymb,amsfonts}
\usepackage{algorithmic}
\usepackage{graphicx}
\usepackage{float}
\usepackage{textcomp}
\usepackage{xcolor}
\usepackage{subfigure}
\usepackage{url}
\usepackage{algorithm}
\usepackage{algorithmic}
\usepackage{makecell}
\usepackage{color}
\usepackage{verbatim}
\usepackage{hyperref}
\usepackage{wrapfig}
\usepackage{multicol}
\usepackage{amsthm}

\def\BibTeX{{\rm B\kern-.05em{\sc i\kern-.025em b}\kern-.08em
    T\kern-.1667em\lower.7ex\hbox{E}\kern-.125emX}}

\begin{document}

\title{Orchestration of Emulator Assisted Mobile Edge Tuning for AI Foundation Models: A Multi-Agent Deep Reinforcement Learning Approach}

\author{\IEEEauthorblockN{Wenhan Yu}
\IEEEauthorblockA{
\textit{Graduate College}\\
\textit{Nanyang Technological University}\\
Singapore \\
wenhan002@e.ntu.edu.sg}\vspace{-10pt}
\and
\IEEEauthorblockN{Terence Jie Chua}
\IEEEauthorblockA{
\textit{Graduate College}\\
\textit{Nanyang Technological University}\\
Singapore \\
terencej001@e.ntu.edu.sg}\vspace{-10pt}
\and
\IEEEauthorblockN{Jun Zhao}
\IEEEauthorblockA{
\textit{School of Computer Science and Engineering}\\
\textit{Nanyang Technological University}\\
Singapore \\
junzhao@ntu.edu.sg}\vspace{-10pt}

}

\maketitle
\pagestyle{plain}
\thispagestyle{plain}

\begin{abstract}
The efficient deployment and fine-tuning of foundation models are pivotal in contemporary artificial intelligence. In this study, we present a groundbreaking paradigm integrating Mobile Edge Computing (MEC) with foundation models, specifically designed to enhance local task performance on user equipment (UE). Central to our approach is the innovative Emulator-Adapter architecture, segmenting the foundation model into two cohesive modules. This design not only conserves computational resources but also ensures adaptability and fine-tuning efficiency for downstream tasks. Additionally, we introduce an advanced resource allocation mechanism that is fine-tuned to the needs of the Emulator-Adapter structure in decentralized settings. To address the challenges presented by this system, we employ a hybrid multi-agent Deep Reinforcement Learning (DRL) strategy, adept at handling mixed discrete-continuous action spaces, ensuring dynamic and optimal resource allocations. Our comprehensive simulations and validations underscore the practical viability of our approach, demonstrating its robustness, efficiency, and scalability. Collectively, this work offers a fresh perspective on deploying foundation models and balancing computational efficiency with task proficiency.
\end{abstract}

\begin{IEEEkeywords}
Mobile edge computing, foundation model, parameter efficient tuning, deep reinforcement learning, wireless communications.
\end{IEEEkeywords}

\section{Introduction} \label{sec:introduction}
\textbf{Background.} 
Artificial intelligence has undergone a profound transformation with the advent of foundation models. These powerful computational structures, like GPT-3~\cite{brown2020language} and BERT~\cite{devlin2018bert}, excel in processing and generating diverse data types, such as text, images, and audio~\cite{foundationmodel}. These models have established new benchmarks in tasks spanning from natural language understanding to content generation and translation. Their strength lies in their extensive training, involving billions of parameters, which fosters a comprehensive and foundational data comprehension. The contemporary era is marked by the dominance of these foundational models, which are increasingly finding applications in various industries such as healthcare, finance, education, and entertainment. These expansive language models offer a versatile toolkit for a wide range of practical applications. They can be tailored or adjusted for specific domains or tasks through a process known as fine-tuning. Fine-tuning~\cite{xiao2023offsite} customizes the initially pretrained model to operate effectively in a more specific context or application. To illustrate, a large language model that has been pretrained on general text data can be fine-tuned to become an expert in tasks like medical diagnosis, legal document review, or customer service chatbots. It's important to recognize that the initial training of foundational models is centered on self-supervised learning from extensive unlabeled data, enabling them to grasp general language comprehension. Conversely, fine-tuning involves adapting these pretrained models to particular tasks using task-specific labeled data, which enhances their specialized performance.

\textbf{Motivation. }Foundation models serve as the cornerstone for a wide array of downstream tasks in industries like finance and healthcare. These models can be customized through fine-tuning to address specific natural language understanding challenges in specialized fields. However, fine-tuning these models for local tasks on mobile devices is prohibitively computationally intensive~\cite{FedPET}. To overcome this challenge, mobile edge computing (MEC) can be employed, where local devices send their training data to a server for model training and fine-tuning. Yet, deploying this solution in decentralized environments, with numerous User Equipments (UEs) handling various tasks, presents complexities. The substantial size of foundation models and local device data, coupled with the high communication and computation costs associated with transmitting data to the server and fine-tuning large language models, pose significant challenges. Moreover, MEC introduces issues like delays in uplink data transmission and model fine-tuning. The key lies in achieving optimal downstream task performance while managing the overhead costs of communication and computation.


\textbf{Proposed solution. }
To address the challenge of fine-tuning foundation models for downstream tasks, we propose a hybrid approach that combines mobile edge computing (MEC) with local device computation. Additionally, we aim to reduce the computational and communication burden of MEC and local device computation by employing an emulator and adapter combination approach~\cite{xiao2023offsite}. An adapter consists of trainable neural network parameters, such as weights, layers, or units, while the emulator is a representation of the fixed-weight portions of the neural network. Adapters enable the server and local devices to train only a subset of the foundation model's parameters, fine-tuning them specifically for the downstream task. Moreover, the emulator, a compressed version of the fixed-weight foundation model, assists in training the adapter during local device model fine-tuning.

To facilitate the adoption of mixed MEC and local device model fine-tuning, we introduce an orchestrator. This orchestrator optimizes crucial variables, including device selection for MEC and local device computation, as well as novel considerations such as emulator compression parameters. The latter is particularly important in the context of foundation model fine-tuning, which was not previously emphasized in MEC approaches. Our orchestrator enhances resource allocation for the fine-tuning process, improving its scalability. Furthermore, our orchestrator leverages a novel multi-agent deep reinforcement learning technique, the Hybrid Multi-agent Proximal Policy Optimization (HMPPO) approach, to seamlessly handle the optimization of both continuous and discrete variables.

Our contributions are as follows:
\begin{itemize}
    \item \textbf{Paradigmatic Shift with MEC}: We introduce a novel paradigm that combines Mobile Edge Computing (MEC) with fine-tuning of foundation models for local device tasks. This approach is designed to enhance model performance for tasks on local user equipment, optimizing computation while maintaining foundation model integrity and performance.
    \item \textbf{Architectural Innovation with Emulator-Adapter}: We divide the foundation model into two components: the Emulator and the Adapter. This modular approach minimizes local device overhead while maintaining foundation model adaptability, achieving a balance between resource conservation and optimizing downstream task-specific model fine-tuning performance.
    \item \textbf{Optimized Resource Allocation Strategy}: We develop an advanced resource allocation mechanism that optimizes key variables, including device selection for mobile edge computing or local device computation, and emulator compression parameters. These variables are selected to address the specific challenges and needs of the Emulator-Adapter structure in a decentralized environment.
    \item \textbf{Hybrid Multi-agent DRL for Resource Allocation}: We deploy a cutting-edge hybrid multi-agent Deep Reinforcement Learning (DRL) method to tackle a mixed discrete-continuous action space problem. This approach effectively addresses the challenges within our system model, enabling optimal and dynamic resource allocation decisions.
    \item \textbf{Comprehensive Simulations and Validations}: We conduct comprehensive simulations and these rigorous tests and evaluations have confirmed the robustness, efficiency, and scalability of our proposed system and solution. These simulations attest to the practical viability and superior performance of our approach.
\end{itemize}

\textbf{Related works. }Foundation models, such as \mbox{GPT-3}\cite{brown2020language} and CLIP\cite{radford2019language}, widely known as large pre-trained models, have gained prominence due to their exceptional ability to make zero-shot predictions and their adaptability to new tasks through a transfer learning method known as fine-tuning~\cite{wei2021finetuned, muennighoff2022crosslingual}. Leveraging these models for fine-tuning to tackle down-stream tasks offers significant advantages in terms of both time and resource savings when compared to the labor-intensive process of training models from scratch.

Efficient utilization of foundation models has become a central focus in modern AI, with techniques designed to reduce computational and storage overhead while maintaining or enhancing performance. Among these techniques, adapters~\cite{houlsby2019parameter} and Low-rank Adapters (LoRA)~\cite{hu2021lora, qlora} have stood out, encoding task-specific information within intermediate layers of a model without overshadowing pre-existing knowledge~\cite{rebuffi2017learning}. The trend in recent advancements, such as Parameter-Efficient Fine Tuning (PEFT), prefix-tuning~\cite{li2021prefix}, and prompt tuning~\cite{qin2021learning,lester2021power}, adapters, P-tuning V2~\cite{liu2021p}, tuning embedding layer inputs~\cite{an2022input}, emphasizes the minimization of changes to model parameters, serving the dual purpose of resource conservation and knowledge encapsulation from larger pre-trained models.

Local devices often face constraints when accommodating the full weight of large foundation models, which not only hinders efficiency but also raises concerns regarding comprehensive model knowledge and potential privacy issues. In response to these challenges, recent research~\cite{xiao2023offsite, ding2023dc} has spotlighted the utility of emulators, scaled-down yet effective versions of foundation models, to facilitate efficient fine-tuning. This approach is particularly relevant within the context of server-assisted computing, where the intertwined future of model tuning and server-assisted computing, specifically within Mobile Edge Computing (MEC), promises resource-sensitive, high-performance AI solutions.

Besides the work by Dong et al.~\cite{dong2023lambo,shen2023large}, there is a notable absence of research dedicated to the mobile edge computing of large foundation models, particularly the implementation of an emulator-assisted approach to fine-tuning such models. Our objective is to pioneer a Parameter-Efficient Emulator Assisted Tuning (PEAT) approach within mobile edge computing to address downstream tasks.

\section{System model} \label{sec:systemmodel}

\begin{figure*}[t] 
\centering
\includegraphics[width=0.9\linewidth]{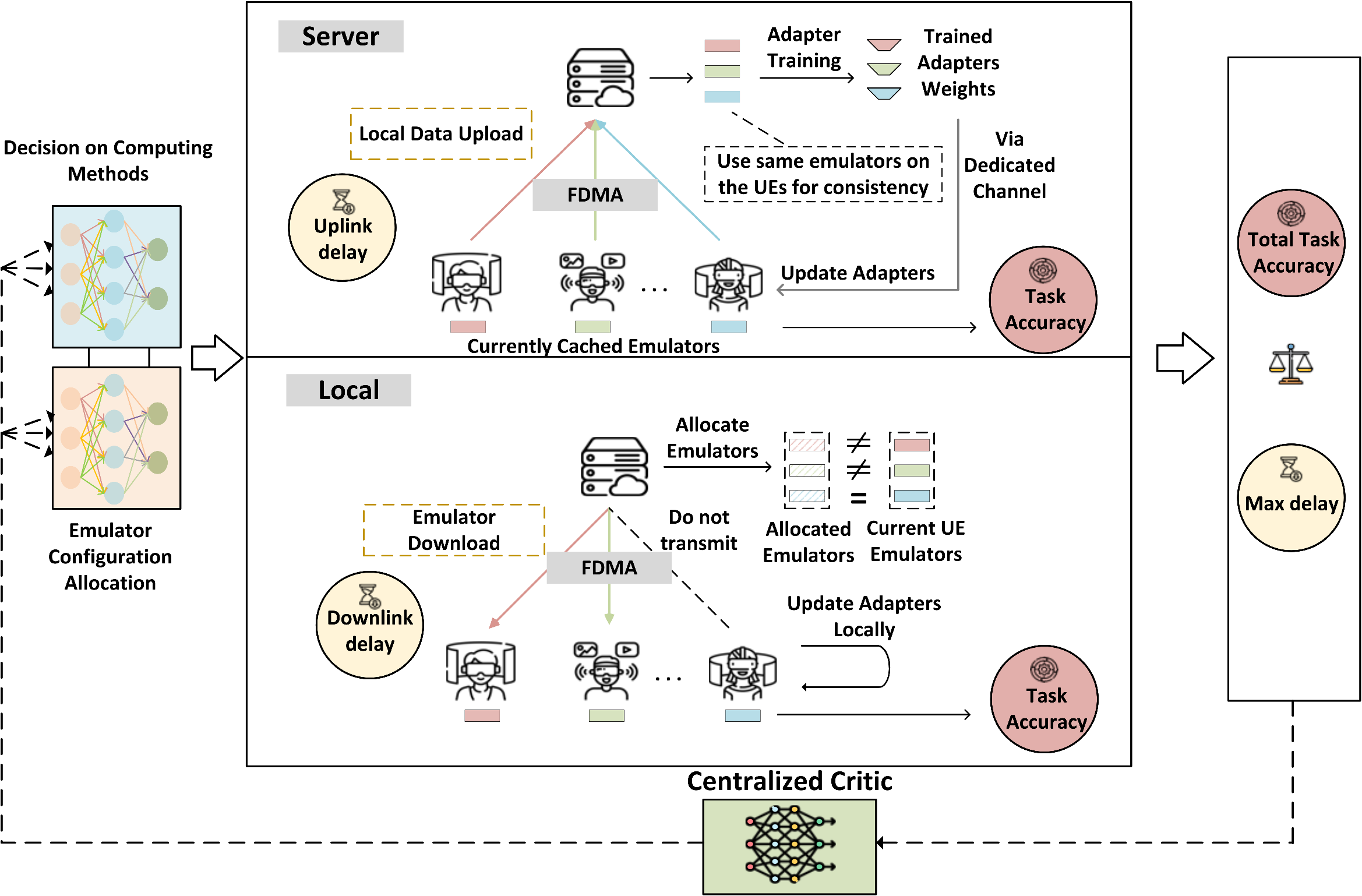}
\caption{Architecture of a central server interacting with User Equipments (UEs) for task execution and resource allocation, where each UE utilizes a two-part model comprising an \textit{Emulator} and an \textit{Adapter}. The decision-making process is governed by Deep Reinforcement Learning, optimizing task accuracy and communication overhead.}
\label{fig:systemmodel}
\end{figure*}

In an environment where a central server operates alongside a collection of User Equipments (UEs) represented by $\mathcal{N}=\{1,2,\dots,N\}$, every UE has a buffer filled with $T_n$ tasks to complete. At every step $t$, each user undertakes a task by utilizing models tailored from the foundational model. These tailored models consist of an \textbf{Emulator}, compressed from the foundation by knowledge distillation, pruning, or layer drop~\cite{modelcompression}, and an \textbf{Adapter} specifically fine-tuned for the task at hand. Assume that each UE can only cache one emulator at a time due to limited storage capacity. Despite this emulator serving as a foundational framework, it doesn't undergo training. Instead, the onus of adaptability lies with the Adapter. With its trainable weights, the Adapter is locally fine-tuned, allowing it to be best aligned with the task it's meant to facilitate.

For our approach, we implement a robust resource allocation method, presuming every UE consistently manages its buffer with a uniform task load, implying $T_n$ remains consistent across all UEs. Addressing this high-demand situation creates a foundation adaptable to cases where some UEs may not have tasks. Central to this architecture, the server maintains a comprehensive foundation model dedicated to bolstering the UEs in their endeavors.

The server, leveraging its advanced algorithms and the foundation model, first determines the (1) optimal emulator configuration $E_n^t$ for each UE ($n$) and the specific task ($t$) it is tackling, beyond this, the server also makes (2) informed decisions $z_n^t$ about where the computation should ideally occur: back at the central server ($z_n^t=1$) or locally at the UE ($z_n^t=0$). Then, the wireless communication overhead and the task accuracy are jointly optimized. The system model is illustrated in Fig.~\ref{fig:systemmodel}. These two cases are expounded on as follows:

\textbf{Case 1: Sever-side Training.} When the UE opts for training to be conducted on the server, the entire process unfolds at the central node. In this scenario, the UE's first step is to securely transmit its data $D_n^t$ to the server. The server, equipped with the necessary computational resources, then engages in the training process by tailoring the foundation model to the $E_n^t$, which is the same as the one currently on the UE designated as $n$, to ensure consistency. Upon the completion of the training phase, the server does not send back the entire model. Instead, it efficiently packages and transmits only the adapter weights. The UE, in turn, employs these weights to update its local adapter, ensuring both the server and UE remain synchronized in their model representations. Thus, the communication overhead for uploading data when using server training is:
\begin{align}
    d_{s,n}^t(z_n^t) = \frac{D_n^t}{r_{u,n}^t} \times z_n^t, ~~\forall n\in\mathcal{N}, \forall t\in\mathcal{T}.
\end{align}
where $r_{u,n}^t$ represents the average upload transmission rate of the $n^{\text{th}}$ UE at task $t$. For transmission, the system employs FDMA to counteract potential interference~\cite{FDMA} and average allocation of bandwidth resources. Consequently, the achievable transmission rate $r_n^t$ can be expressed as:
\begin{align}
    r_{u,n}^t(z_n^t) = \Bar{W}_u^t \log_2(1+\frac{p_{u,n}^t h_n^t}{\sigma^2\Bar{W}_u^t}), \label{uprate} 
\end{align}

where $p_{u,n}^t$ is the uplink transmission power of UE $n$ determined by the respective device, and $\boldsymbol{h}_n^t$ is the average channel gain when dealing with task $t$, which will be expounded on in Section~\ref{sec:numerical}. $\Bar{W}_u^t$ is the average uplink bandwidth:
\begin{align}
    \Bar{W}_u^t(z_n^t) = \frac{W_u^s}{\{|\mathcal{N}|^t\}_{\forall n:z_n^t=1}}.
\end{align}
Here, the $\{|\mathcal{N}|^t\}_{\forall n:z_n^t=1}$ is the number of UEs allocated on server training, and $W_u^{s}$ is the sum resource of the uplink bandwidth.

In Case 1, we bypass the downlink overhead for transmitting adapter weights. This is justified by their minimal size compared to the emulators, as corroborated by \cite{xiao2023offsite}. Furthermore, we rely on dedicated channels for this transmission, ensuring efficiency.

\textbf{Case 2: Local Training.} Alternatively, if the UE decides to manage its training locally, the server assumes a consultative role. It reviews the UE's current emulator and, if deemed unsuitable for the present task compared to the UE's previous emulator $E_n^{t-1}$, the server dispatches an updated emulator to UE $n$. Essentially, if the cached emulator on the UE suffices for the subsequent task, the server can endorse the continued use of the same emulator, thereby saving on transmission overhead. We introduce an auxiliary emulator switch indicator $I_n^t$ to capture this:

\noindent For $\forall n\in\mathcal{N}, \forall t\in\mathcal{T}, \text{ and } z_n^t=0$:
\begin{align}
    I_n^t \hspace{-2pt}=\hspace{-2pt}
    \begin{cases}
        0, \text{ if } E_n^t = E_n^{t-1}.\\
        1, \text{ otherwise}.
    \end{cases}
\end{align}

Then the communication overhead is:
\begin{align}
    d_{l,n}^t(z_n^t, E_n^t) \hspace{-2pt}=\hspace{-2pt} \frac{E_n^t\times D_{FM}}{r_{d,n}^t} \hspace{-2pt}\times\hspace{-2pt} I_n^t \hspace{-2pt}\times\hspace{-2pt} (1\hspace{-2pt}-\hspace{-2pt}z_n^t), \forall n\in\mathcal{N}, \forall t\in\mathcal{T},
\end{align}
where the $D_{FM}$ is the size of the original foundation model and downlink rate $r_{d,n}^t$ is:
\begin{align}
    r_n^t(z_n^t, E_n^t) = \Bar{W}^t \log_2(1+\frac{\Bar{p}^t h_n^t}{\sigma^2\Bar{W}^t}), \label{downrate} 
\end{align}

For both bandwidth and power, average allocations of the aggregate resources on the server are utilized as:
\begin{align}
    &\Bar{W}^t(z_n^t, E_n^t) = \frac{W_d^{s}}{\{|\mathcal{N}|^t\}_{\forall n:(z_n^t=1\text{ and }I_n^t=1)}}, \nonumber \\
    &\Bar{p}^t(z_n^t, E_n^t) = \frac{P_d^{s}}{\{|\mathcal{N}|^t\}_{\forall n:(z_n^t=1\text{ and }I_n^t=1)}}.
\end{align}

In Case 2, the proactive approach ensures that the UE always operates with the most appropriate version of the emulator for its tasks. Once equipped with the correct emulator, the UE takes the reins, conducting the training of the adapter on its own. This localized approach eliminates the need for any data uploads to the server but may require additional communication resources for downloading the new emulator.

For the communication overhead, in each step $t$, the uplink (data transmission for server computing) and downlink (emulator transmission for local computing) happen simultaneously, and the maximum delay of all users at $t$ is taken as the system latency. Therefore, the delay for each user $d_n^t$ can be shown as:
\begin{align}
    d_n^t(z_n^t,E_n^t) = d_{s,n}^t\times z_n^t + d_{l,n}^t\times I_n^t \times (1-z_n^t). \label{eq:userdelay}
\end{align}

\textbf{Task accuracy.} In the real-world scenario, tasks naturally differ in their demands and intricacies. To address this and to bring about a more accurate representation of the situation, we introduce a quantified factor $c_n^t$. This factor serves as a metric, capturing the inherent complexity of each task. By quantifying the complexity, the system can make more informed decisions, allowing for better allocation of resources and more precise emulator configurations. Thus, the task accuracy is formulated as:
\begin{align}
    \kappa_n^t(z_n^t, E_n^t) = 
    \begin{cases}
        f(1) \times \frac{\iota}{c_n^t+\iota}, \text{ if } z_n^t=1.\\
        f(E_n^t)\times \frac{\iota}{c_n^t+\iota}, \text{ otherwise}. \label{eq:accuracy}
    \end{cases}
\end{align}
Here, $E_n^t$ represents the layer drop retention accuracy less than $1$. By treating layer retention as a continuous variable, the system gains fine-grained control over the emulator's size, allowing for a dynamic balance between accuracy and communication overhead. The $f(\cdot)$ captures the accuracy based primarily on $E_n^t$, which is curve-fitted based on empirical results from~\cite{xiao2023offsite}, further detailed in Section~\ref{sec:numerical}. The term $\frac{\iota}{c_n^t+\iota}$ serves to scale down $\kappa_n^t$ as $c_n^t$ increases. As $c_n^t$ grows, the scaling factor diminishes, thus reducing $\kappa_n^t$.

\section{Problem Formulation}
The crux of our problem lies in striking an optimal balance between model accuracy and communication overhead. This trade-off is sought by dynamically dictating two primary factors: the computing cases (either on the server or locally) and the update frequency and emulator layer drop retention.

To systematize this dynamic allocation, we employ two matrices, $\boldsymbol{z}$ and 
$\boldsymbol{E}$. These matrices capture the allocation patterns for both computing cases and emulator configurations across all users and tasks. Specifically, the entry at the $n^{\text{th}}$ row and $t^{\text{th}}$ column of these matrices corresponds to $z_n^t$ and $E_n^t$, respectively.

Our primary objective, with respect to task accuracy, is to enhance the cumulative accuracy across all tasks:
\begin{align}
    \text{P1: } \max_{\boldsymbol{z},\boldsymbol{E}}\sum_{t=1}^T\sum_{n=1}^{N} \kappa_n^t. \label{eq:accuracyobj}
\end{align}

Concurrently, we are also concerned with the overall communication overhead, aiming to minimize it:
\begin{align}
    \text{P2: } \min_{\boldsymbol{z},\boldsymbol{E}}\sum_{t=1}^T \max_{n\in\mathcal{N}}(d_n^t). \label{eq:latencyobj}
\end{align}

Synthesizing these objectives, the overarching problem encapsulating both Eq.~(\ref{eq:accuracyobj}) and Eq.~(\ref{eq:latencyobj}) can be concisely formulated as:
\begin{align}
    &\max_{\boldsymbol{z},\boldsymbol{E}}\left\{ \omega_1\times \sum_{t=1}^T\sum_{n=1}^{N} \kappa_n^t - \omega_2 \times \sum_{t=1}^T \max_{n\in\mathcal{N}}(d_n^t) \right\}, \label{eq:problemformulation} \\
    & s.t.~C1: z_n^t \in \{0,1\}, \forall n\in\mathcal{N}, \forall t\in\mathcal{T}.\\
    &~~~~~C2: 0\leq E_n^t \leq 1, \forall n\in\mathcal{N}, \forall t\in\mathcal{T}.
\end{align}
where the $\omega_1, \omega_2$ are weight parameters, derived from specific reward settings which are expounded upon in Section~\ref{sec:environment}.

The Emulator-Adapter framework provides flexibility, ensuring that while communication costs are optimized, the accuracy is not compromised. The system recognizes that not all tasks are created equal, and its design caters to these differences, striking a balance between efficient resource use and effective task completion. The formulated problem in 
Eq.~(\ref{eq:problemformulation}) contains highly coupled both discrete (computing cases decisions) and continuous (emulator configuration), making it a tough inseparable mixed integer non-linear programming (MINLP) problem and the sequential nature of this problem further complicates the problem. Thus, it is infeasible to use traditional optimization strategies, and Deep Reinforcement Learning (DRL) algorithms, attributed to their superior ability to tackle sequential problems and find near-optimal solutions, need to be considered.

\section{DRL Environment Setting} \label{sec:environment}
At present, model-free DRL methods are well utilized in wireless communication scenarios~\cite{wirelessDRL}, since they can efficiently reach near-optimal points while tackling a number of random and unpredicted factors. For model-free DRL, three key elements are essential to create the DRL environment based on the formulated problem, allowing agents to interact with and learn satisfactory policies. Thereafter, we provide the detailed settings of these three elements: state, action, and reward.

\subsection{State}
Since the computing case decisions and emulator configuration need to be jointly optimized, involving mixed discrete-continuous actions, we propose to use two agents for optimizing them.

\textbf{Agent 1 (case decisions):} The state of Agent 1 $s_1^t$ includes (1) average channel gains of currently finished tasks of all users $\{h_1^{t-1}, h_2^{t-1},\dots, h_N^{t-1}\}$. (2) the current task complexities of all users $\{c_1^t,c_2^t,\dots,c_N^t\}$. (3) current local data sizes of all users $\{D_1^t, D_2^t, \dots, D_N^t\}$. (4) the currently cached emulators on different UEs $\{E_1^t, E_2^t,\dots,E_N^t\}$.

\textbf{Agent 2 (emulator configuration):} The state of Agent $s_2^t$ involves: (1) the currently cached emulators on different UEs like in Agent 1. (2) the action $a_1^t$ from Agent 1 (cases decisions). (3) the task complexities of all users as in Agent 1.

\subsection{Action}
The actions are intuitive. For Agent 1 controlling the case decisions, the discrete action contains all decisions for different users $\{z_1^t,z_2^t,\dots,z_N^t\}$, and in terms of Agent 2 handling the emulator configuration allocation, continuous actions $\{E_1^t,E_2^t,\dots,E_N^t\}$ is used.

\subsection{Reward}
Utilizing the Centralized Training Decentralized Execution (CTDE) framework, the global reward is set to give feedback to the Critic and learn the state value, then update the two Actors. This reward $R_g^t$ is composed of (1) the average task accuracy among users of the current step: $\omega_p \times\frac{1}{N}\times\sum_{n\in\mathcal{N}}\kappa_n^t$. In the simulation, the accuracy refers to the language model perplexity, lower is preferable, further detailed in Section~\ref{sec:numerical}. Thus, the weight is set as negative. (2) the maximum communication delay at $t$: $\omega_d\times\max_{n\in\mathcal{N}}(d_n^t)$, where the $\omega_p, \omega_d$ are negative weight parameters.

\section{Methodology} \label{sec:methodology}
\subsection{Preliminary}
Proximal Policy Optimization (PPO) by OpenAI~\cite{PPO} offers significant advancements over traditional policy gradient algorithms. PPO's strengths can be attributed to its enhanced sample efficiency and the introduction of a policy constraint.

\begin{enumerate}
\item \textit{Sample Efficiency Enhancement:}
PPO uses two distinct policies: $\pi_{\theta'}$ for sampling trajectories during training, and $\pi_\theta$ for evaluation. This separation optimizes the algorithm's sample efficiency. The expectation relationship between them is expressed as:
\begin{align}
    \mathbb{E}_{(s^t,a^t)\sim\pi_\theta}[\pi_\theta(a^t|s^t) A^t] &= \mathbb{E}_{(s^t,a^t)\sim\pi_{\theta^{'}}}[\frac{\pi_\theta(a^t|s^t)}{\pi_{\theta^{'}}(a^t|s^t)} A^t]. \label{eq:sample}
\end{align}
where $A^t$ is the advantage function to estimate how is the selected action.

\item \textit{Introduction of Policy Constraint:}
Switching between the data sampling policies doesn't eliminate variances between their objective functions. To tackle this, a KL-divergence penalty is integrated into the reward formulation. Due to the impracticality of computing the KL divergence for every observation, the objective function is redefined as~\cite{PPO}:
\begin{align}
    \mathbb{E}_{(s^t,a^t)\sim\pi_{\theta_{'}}}[f^t(\theta)A^t],
\end{align}
where
\begin{align*}
    f^t(\theta)=\min\{r^t(\theta), \text{clip}(r^t(\theta), 1-\epsilon, 1+\epsilon)\}.
\end{align*}
Here, $r^t(\theta)$ represents the ratio between the two policies: $r^t(\theta)\hspace{-2pt}=\hspace{-2pt}\frac{\pi_\theta(a^t|s^t)}{\pi_{{\theta'}}(a^t|s^t)}$. And the advantage function $A^t$ is calculated via Generalized Advantage Estimation (GAE)~\cite{GAE}:
\begin{align}
    &A^t = \delta^t + (\gamma\lambda)\delta^{t+1}+...+(\gamma\lambda)^{\bar{T}-1}\delta^{t+\bar{T}-1}, \\
    &\text{where}~~~\delta^t=R^t+\gamma V_{\phi'}(s^{t+1})-V_{\phi'}(s^t).
\end{align}

The gradient associated with this problem is captured by:
\begin{align}
    \Delta\theta = \mathbb{E}_{(s^t,a^t)\sim\pi_{\theta_{'}}}[\triangledown f^t(\theta)A^t]. \label{eq:actorobj}
\end{align}

\item \textit{Value Network (Critic) Implementation:}
PPO employs a Critic reminiscent of other Actor-Critic algorithms. The loss function is defined as:
\begin{align}
    L(\phi) = [V_\phi(s^t)-(A^t+V_{\phi'}(s^{t}))]^2. \label{eq:criticloss}
\end{align}
In this context, $V(s)$ is the widely-recognized state-value function~\cite{RLintro}. With $\phi$ being the learned parameter, it's updated by minimizing $L(\phi)$. The target state-value function, parameterized by $\phi'$, is periodically updated in alignment with $\phi'$. This approach of using a target value is a staple in RL, a strategy embedded in numerous algorithms~\cite{RLintro}.
\end{enumerate}

\subsection{HMPPO}
In this study, we introduce the Hybrid Multi-agent PPO (HMPPO), a specialized variant of the multi-agent PPO (MAPPO) algorithm tailored for both discrete and continuous actions. Notably, the intrinsic design of the PPO algorithm emphasizes evaluating state values (a.k.a. V value) over action values (a.k.a. Q value), which is beneficial as it simplifies the learning process for agents, as evidenced in~\cite{liyueheng}. Consequently, in the PPO structure, the action does not form part of the Critic's input, equipping PPO to seamlessly cater to both discrete and continuous action sets.

Building on the principles of the CTDE paradigm, as presented in~\cite{CTDE}, and integrating the hybrid structure for addressing mixed actions, we elucidate the update functions associated with the two Actors and a single Critic in our proposed HMPPO framework:
\begin{align}
    &\Delta\theta_1 \hspace{-2pt}=\hspace{-2pt} \mathbb{E}_1^t[\nabla_{\theta_1} \min\{r^t(\theta_1)A^t, \text{clip}(r^t(\theta_1), 1\hspace{-2pt}-\hspace{-2pt}\epsilon, 1\hspace{-2pt}+\hspace{-2pt}\epsilon)A^t\}], \nonumber\\
    &\Delta\theta_2 \hspace{-2pt}=\hspace{-2pt} \mathbb{E}_2^t[\nabla_{\theta_2} \min\{r^t(\theta_2)A^t, \text{clip}(r^t(\theta_2), 1\hspace{-2pt}-\hspace{-2pt}\epsilon, 1\hspace{-2pt}+\hspace{-2pt}\epsilon)A^t\}], \label{eq:actor}\\
    &L^t(\phi) = [V_\phi(\{s_1^t;s_2^t\})-(A^t+V_{\phi'}(\{s_1^t;s_2^t\}))]^2, \label{eq:critic}
\end{align}
where the $\{s_1^t;s_2^t\}$ is the concatenation of these two states.

\begin{figure*}[t]
\centering
\subfigtopskip=2pt
\subfigbottomskip=2pt

\subfigure[Training reward with 8 UEs.]{
\begin{minipage}[t]{0.33\linewidth}
\centering
\includegraphics[width=1\linewidth]{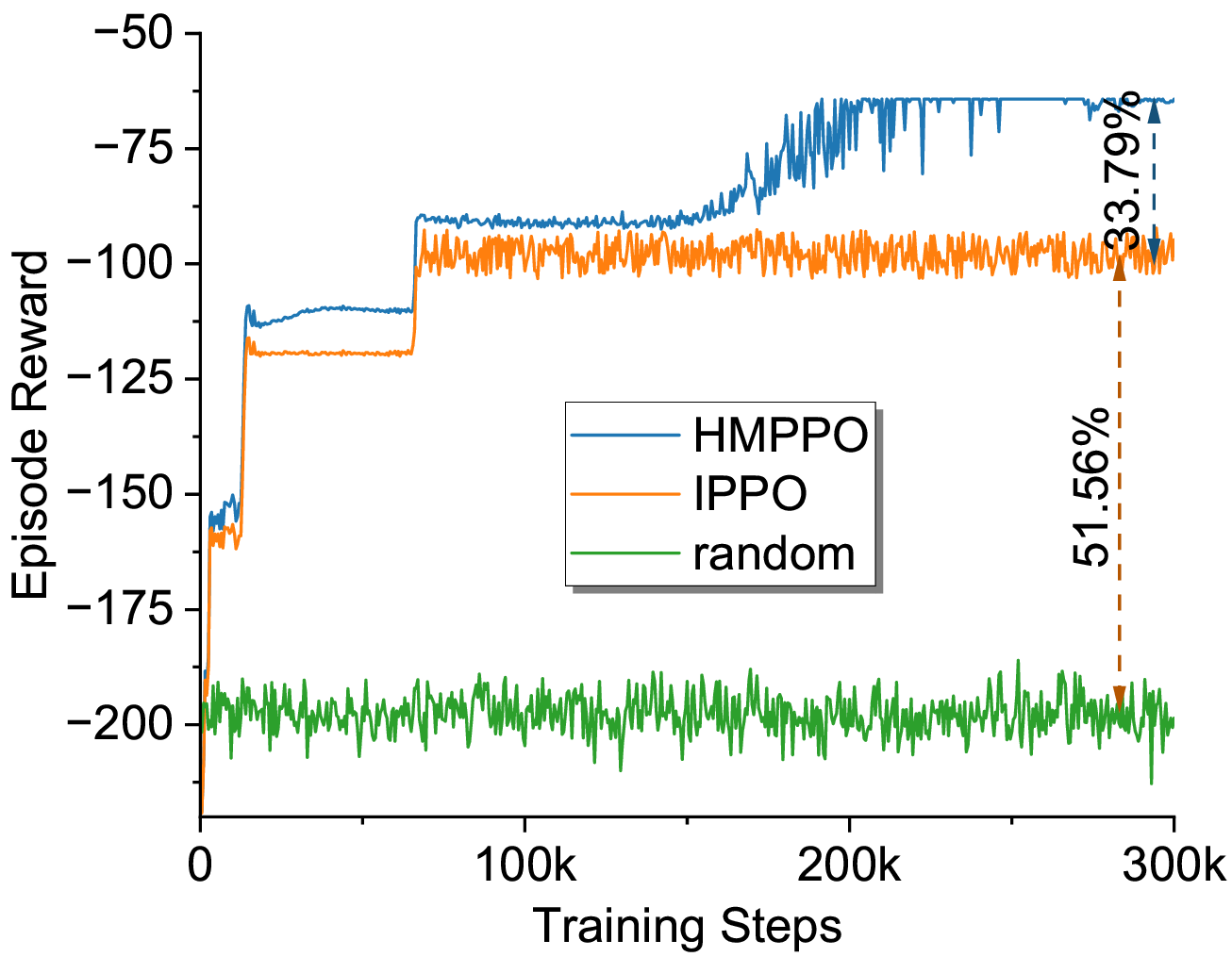}
\label{fig:trainreward}
\vspace{-10mm}
\end{minipage}
}%
\subfigure[Total delay of all tasks (minutes) with 8 UEs.]{
\begin{minipage}[t]{0.33\linewidth}
\centering
\includegraphics[width=1\linewidth]{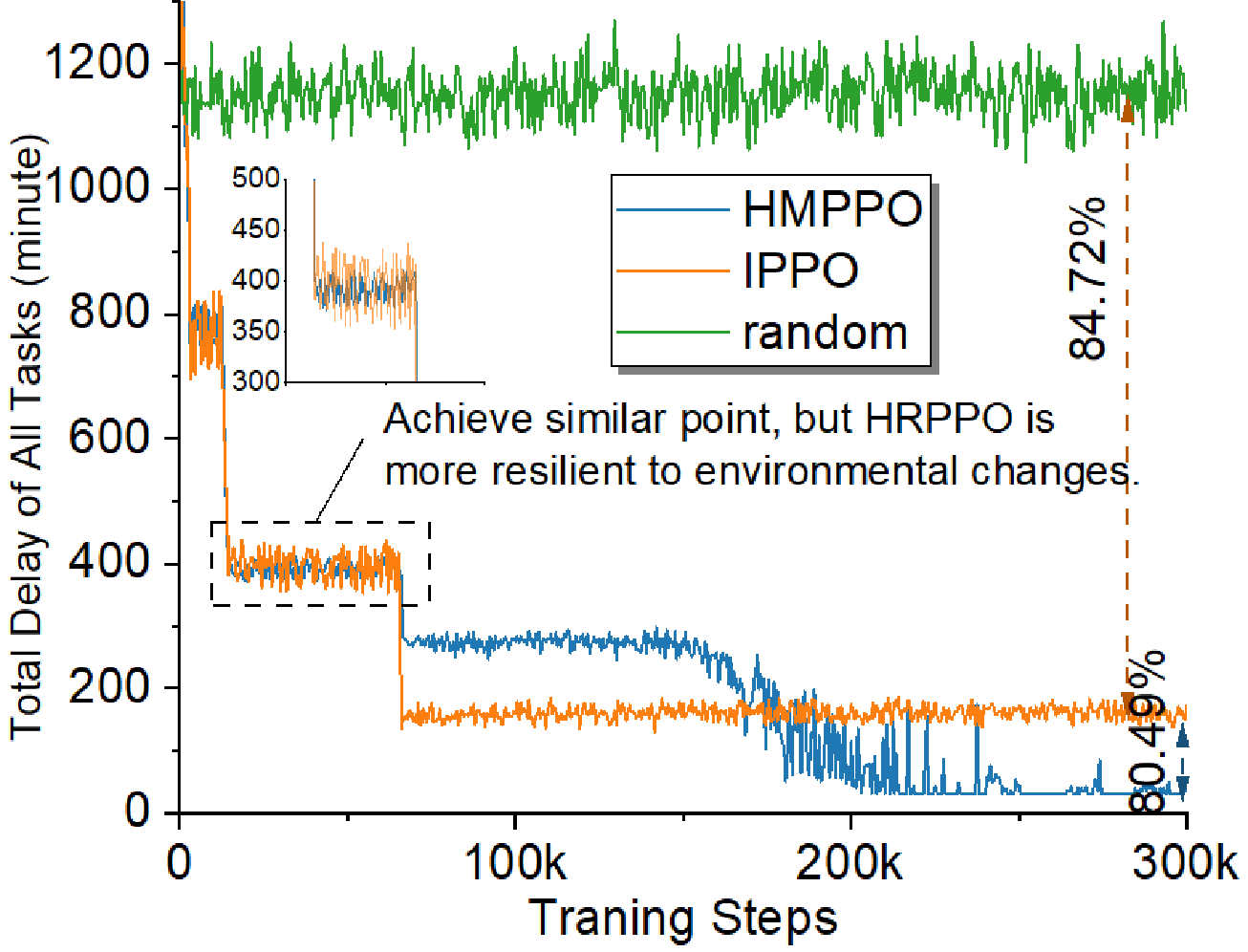}
\label{fig:traindelay}
\vspace{-10mm}
\end{minipage}
}%
\subfigure[Average task perplexity with 8 UEs.]{
\begin{minipage}[t]{0.33\linewidth}
\centering
\includegraphics[width=1\linewidth]{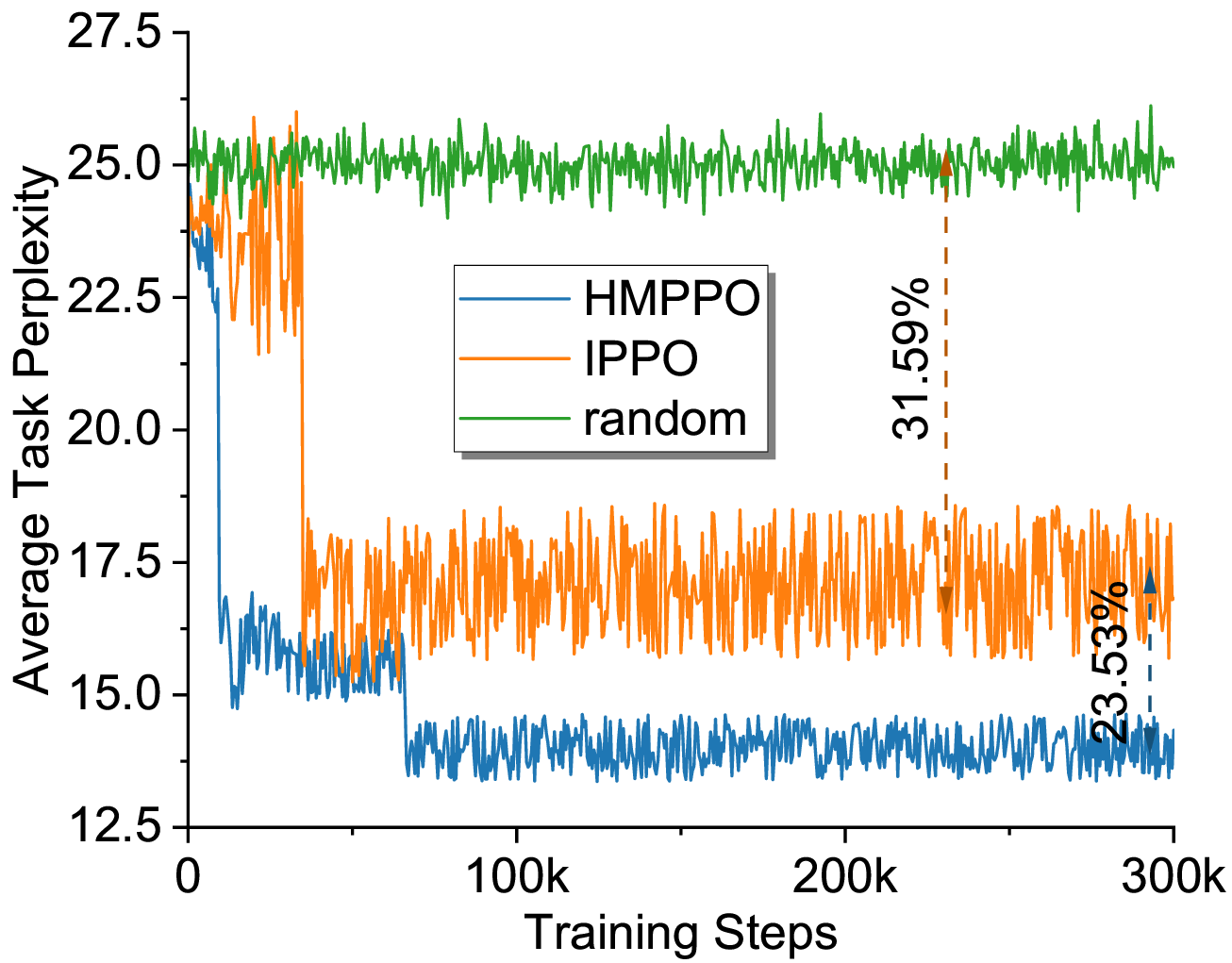}
\label{fig:trainperplexity}
\vspace{-10mm}
\end{minipage}
}%

\subfigure[Emulator change times with 8 UEs.]{
\begin{minipage}[t]{0.33\linewidth}
\centering
\includegraphics[width=1\linewidth]{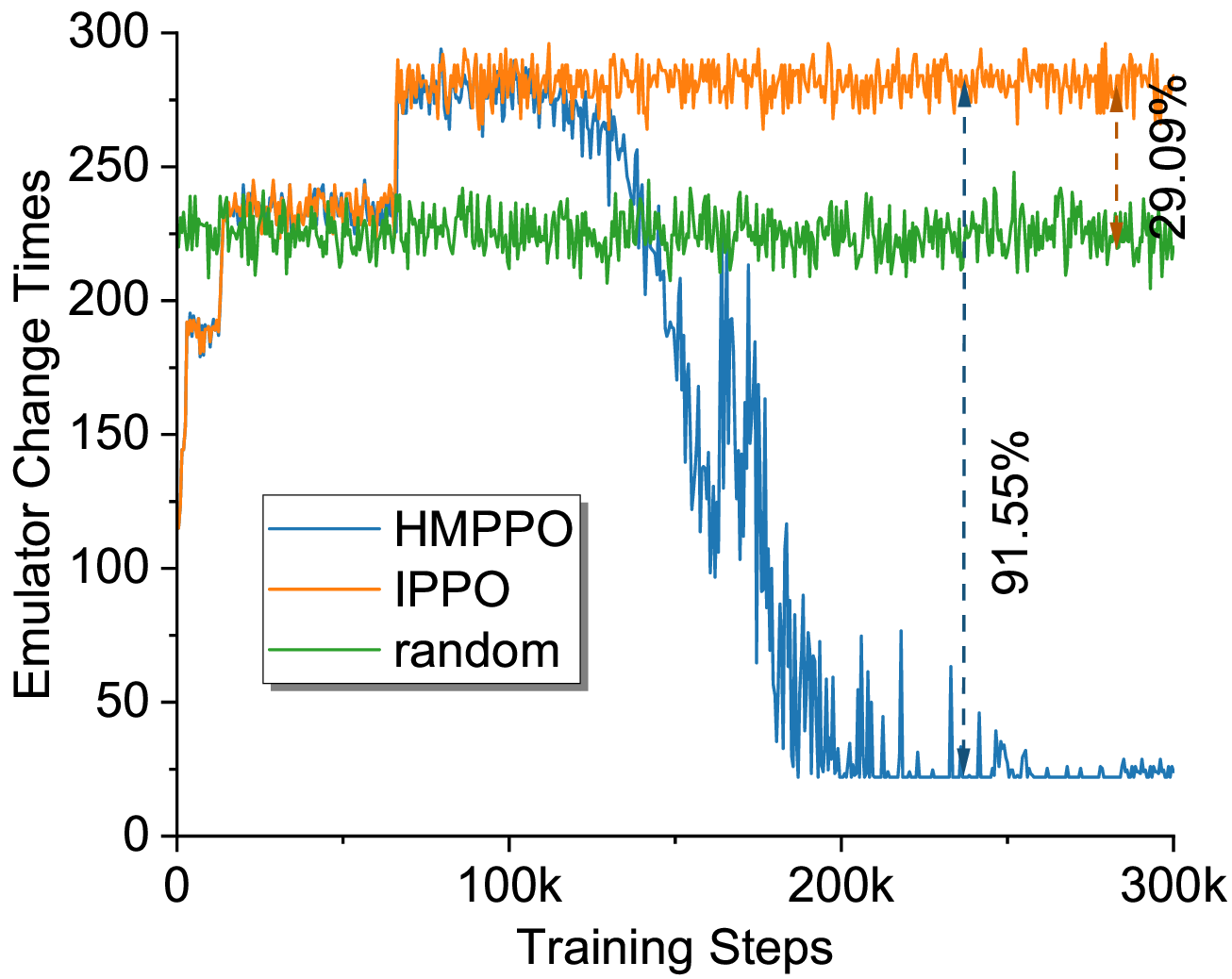}
\label{fig:trainemulator}
\vspace{-10mm}
\end{minipage}
}%
\subfigure[Total delay with 6 to 9 UEs (minutes).]{
\begin{minipage}[t]{0.33\linewidth}
\centering
\includegraphics[width=1\linewidth]{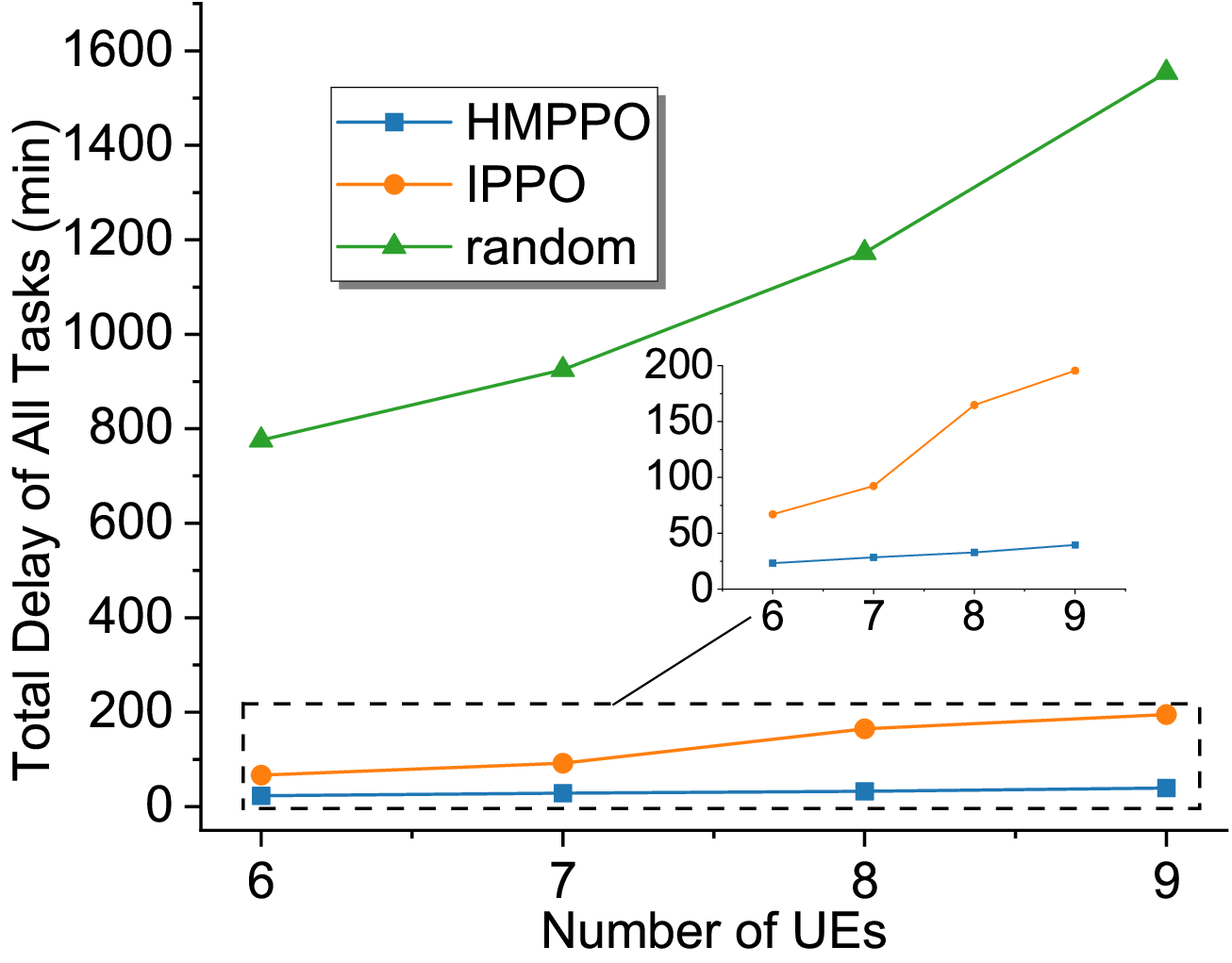}
\label{fig:delayall}
\vspace{-10mm}
\end{minipage}
}%
\subfigure[Average task perplexity with 6 to 9 UEs.]{
\begin{minipage}[t]{0.33\linewidth}
\centering
\includegraphics[width=1\linewidth]{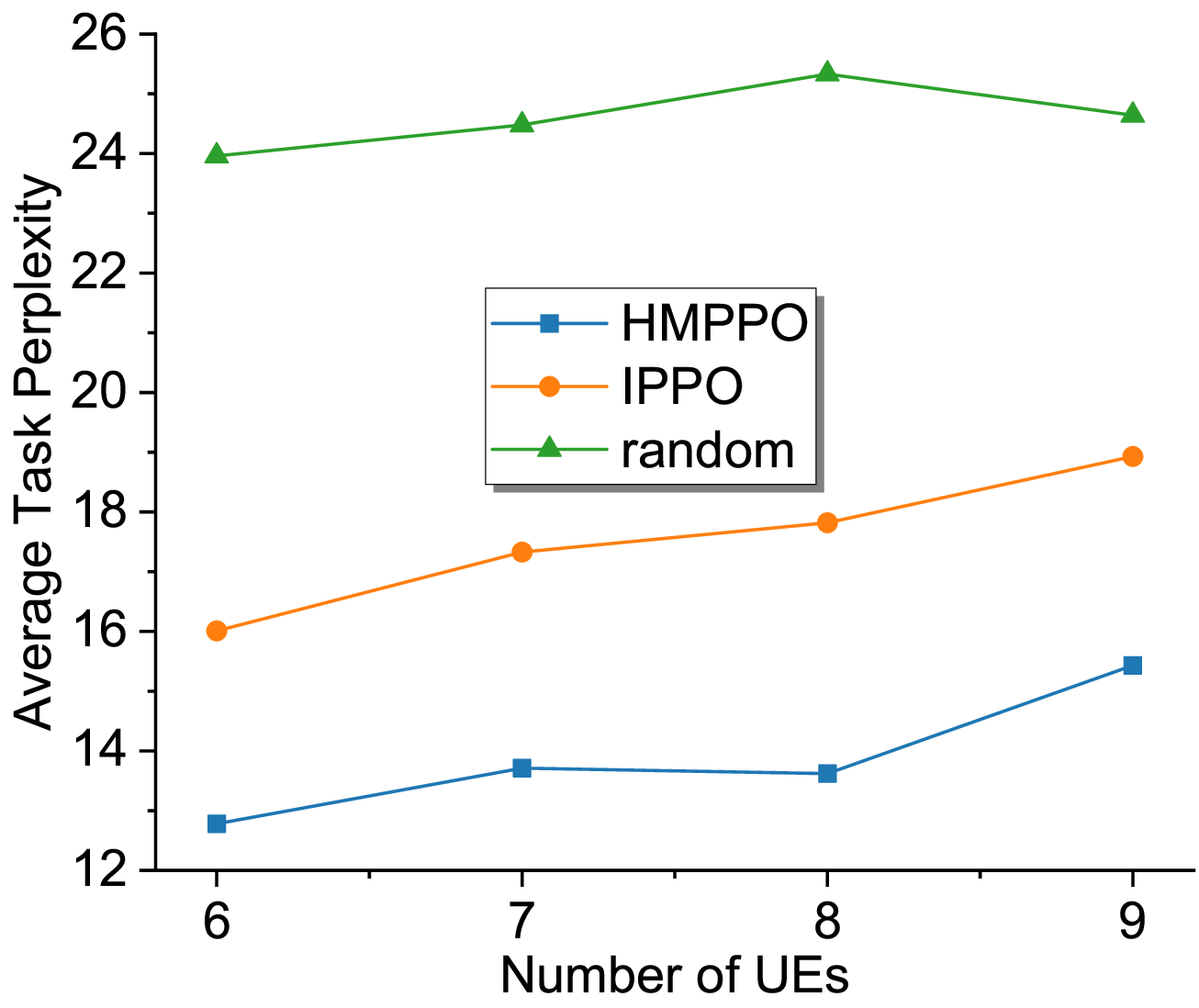}
\label{fig:perplexityall}
\vspace{-10mm}
\end{minipage}
}%

\caption{Simulation results. The first four sub-figures illustrate complete training results on different metrics (i.e., rewards, system delay, task perplexity, and emulator switch times), and the final two sub-figures delineate the overall results on delay and perplexity across all scenarios, where the number of UEs varies from $6$ to $9$.}
\label{fig:train}
\vspace{-0.5cm}
\end{figure*}

\section{Simulations} \label{sec:simulation}
\subsection{Numerical Settings} \label{sec:numerical}
We configured the number of UEs to range from $6$ to $9$ across various experimental setups. The foundation model has a size of $10.8$\,GB, which corresponds to the \textit{GPT-3 2.7B} large language model~\cite{size}. The number of tasks $T$ for each UE is set to $50$. The emulator retention is adjusted between $0.2$ and $0.8$. Local data sizes, denoted as $D_n^t$, are randomly chosen from a uniform distribution spanning $300$\,MB to $500$\,MB. The upload transmission power is uniformly selected from a range of $200$\,mW to $1$\,W. Assuming the use of Frequency Division Duplexing (FDD) to allocate distinct bandwidth resources for uplink and downlink transmissions~\cite{FDD}, the aggregate bandwidth limits are set to $10^5$\,Hz for uplink and $10^6$\,Hz for downlink. With regard to channel gain, we assume the channel remains coherent over short intervals of $10$\,ms. Small-scale fading adheres to the Rayleigh distribution, with a path loss exponent of $\alpha = 2$. The total power resource allocated for downlink is $60$\,W. 

For the accuracy function $f(E_n^t)$ in $\kappa_n^t$, we performed curve-fitting on the relationship between layer-drop-retention and large language model (LLM) perplexity from~\cite{xiao2023offsite}, yielding the function $f = 25.2(E_n^t)^2 - 43.1E_n^t + 31.9$ with an $R^2$ score of $0.97$, indicating a high level of fit to the observed data. And then $\kappa_n^t=f(E_n^t)\times \frac{c_n^t+10}{c_n^t}$, where $\iota=10$ is the complexity weight parameter. Note that we use $\frac{c_n^t+\iota}{c_n^t}$ instead of $\frac{\iota}{\iota+c_n^t}$ in Eq.~(\ref{eq:accuracy}) since we use perplexity and lower is better. Perplexity is a measure used in language modeling to quantify how well a probability distribution predicts a sample~\cite{xiao2023offsite}, a lower perplexity indicates the model's predictions are closer to the true distribution. It's important to note that in the context of our simulation, task accuracy is represented by the LLM perplexity, where \textbf{lower values are preferable}. All experiments were conducted using a single NVIDIA GTX 2080 Ti. We employed $3\times 10^5$ training steps, with evaluation intervals set at every $500$ training steps, and all experiments are conducted under the same global random seed.

\subsection{Metrics and baselines}
The most important metrics in the paper are:
\begin{enumerate}
    \item The DRL episodic reward is the direct feedback for the agents, and it serves as intuitive evidence for comparing algorithms' performances.
    \item The average perplexity among all tasks of all users, showcases our task accuracy.
    \item The total communication delay of all UEs across $T$ tasks, testifies to the communication overhead.
    \item Other than the two above-mentioned matrices corresponding to the reward composition in Section~\ref{sec:environment}, we also include emulator change times, to show if the algorithms catch the sequential nature and try to reduce the emulator changes to decrease the communication overhead for downloading.
\end{enumerate}

We design the following baseline algorithms to compare with our proposed HMPPO:
\begin{itemize}
    \item \textbf{Independent PPO (IPPO)~\cite{IPPO}}: A straightforward approach to utilizing RL in a cooperative interactive setting involves deploying two independent RL agents that interact with each other. We employed this concept using two independent PPO agents.
    \item \textbf{Random}: This method involves two agents selecting actions at random, representing system performance without any optimization strategy. The random policy acts as a baseline, showcasing results in the absence of optimization.
\end{itemize}

\section{Result analysis}
As illustrated in Fig.~\ref{fig:train}, the performance of the HMPPO method consistently surpasses both IPPO and the random approach across all metrics, which includes reward, delay, perplexity, and emulator changes. During the initial stages, HMPPO and IPPO exhibit comparable performance, even when pitted with the total delay, IPPO momentarily outshines HMPPO in system delay around the $100$\,k steps mark, as shown in Fig.~\ref{fig:traindelay}. However, HMPPO eventually demonstrates marked improvement across the board. To provide specific figures: HMPPO exceeds IPPO by margins of $33.79\%$, $80.49\%$, $23.53\%$, and $91.55\%$ for reward, delay, perplexity, and emulator changes, respectively. It's worth noting that IPPO maintains a consistent lead over the random approach. Since total delay and task perplexity directly influence the reward, the patterns observed in them closely align with each other. Yet, the emulator change times metric reveals distinct and interesting behaviors. While both HMPPO and IPPO initially increase emulator change times to better adapt and enhance the total reward, only HMPPO eventually optimizes by decreasing emulator change times around $150$\,k steps. This optimization benefits by reducing the total delay, as evident in Fig.\ref{fig:traindelay}, which subsequently contributes to the reward boost shown in Fig.\ref{fig:trainreward}. An essential observation is the resilience of HMPPO compared to IPPO. Despite achieving similar peaks during the initial stages, IPPO exhibits considerable fluctuations in performance, as highlighted in Fig.\ref{fig:traindelay} and Fig.\ref{fig:trainperplexity}, suggesting that HMPPO provides a more stable and reliable optimization approach.

From Table.~\ref{table:results}, it's evident that the HMPPO method consistently outperforms both IPPO and the random approach across all UE numbers in terms of reward, total delay, and task perplexity. As the UE number increases, while all methods exhibit a decrease in reward and an increase in both total delay and task perplexity, HMPPO's degradation is much more gradual, highlighting its scalability and robustness. The delay analysis further showcases HMPPO's efficiency, with the method maintaining a reasonably low total delay even with an increase in UE numbers, whereas IPPO's delay increases substantially, especially when transitioning from 7 to 8 UEs. On the aspect of task perplexity, HMPPO again emerges superior, offering the lowest values across the board, ensuring a more precise task-specific understanding. This is in stark contrast to the random method, which exhibits the highest perplexity, reflecting its general inefficiency. Lastly, in terms of reward, HMPPO achieves the least negative values consistently, underscoring its performance advantage. Overall, the trends in the table affirm HMPPO's effectiveness and scalability, making it an optimal choice in environments with varying UE numbers. 

\begin{table}[t]
\centering
\caption{Overall results}
\label{table:parameter}
\vspace{-0.2mm}
\scalebox{1}{
\begin{tabular}{ccccc}
\hline
UE number & Reward & \makecell{Total delay (min)} & \makecell{Task perplexity} \\ \hline
\multicolumn{4}{c}{HMPPO} \\ \hline
$6$ & $-57.32$ & $23.25$ & $12.78$ \\
$7$ & $-62.74$ & $28.53$ & $13.71$ \\
$8$ & $-64.23$ & $32.83$ & $13.62$ \\
$9$ & $-67.42$ & $39.64$ & $15.43$ \\ \hline
\multicolumn{4}{c}{IPPO} \\ \hline
$6$ & $-78.98$ & $66.88$ & $16.01$ \\
$7$ & $-86.35$ & $92.33$ & $17.33$  \\
$8$ & $-95.52$ & $164.73$ & $17.82$  \\
$9$ & $-113.74$ & $195.32$ & $18.93$\\ \hline
\multicolumn{4}{c}{random} \\ \hline
$6$ & $-175.57$ & $775.97$ & $23.96$ \\
$7$ & $-193.40$ & $924.89$ & $24.477$ \\
$8$ & $-200.63$ & $1172.45$ & $25.33$ \\
$9$ & $-240.11$ & $1553.73$ & $24.64$ \\ \hline
\end{tabular}
}
\label{table:results}

\end{table}

\section{Conclusion}
Throughout this study, we pioneered an innovative approach by combining Mobile Edge Computing (MEC) with the fine-tuning of foundation models to optimize local device tasks. The introduction of our Emulator-Adapter framework is a testament to our commitment to optimizing performance without overburdening device resources. Our innovative resource allocation strategy ensures that our system thrives in a decentralized setting. Our results, particularly with the HMPPO method, underline the efficacy of our approach. With remarkable improvements across key metrics, like a $33.79\%$ enhancement in reward when compared to IPPO, our approach stands validated. The application of a hybrid multi-agent Deep Reinforcement Learning (DRL) further confirmed the resilience and adaptability of our system model.

In summation, our research offers a promising pathway in the realm of AI, particularly for the deployment of extensive machine learning models on everyday devices. While we have made significant strides, the horizon is vast, and we anticipate even more refined and efficient solutions in the future.

\bibliographystyle{IEEEtran}

\begin{thebibliography}{10}
\providecommand{\url}[1]{#1}
\csname url@samestyle\endcsname
\providecommand{\newblock}{\relax}
\providecommand{\bibinfo}[2]{#2}
\providecommand{\BIBentrySTDinterwordspacing}{\spaceskip=0pt\relax}
\providecommand{\BIBentryALTinterwordstretchfactor}{4}
\providecommand{\BIBentryALTinterwordspacing}{\spaceskip=\fontdimen2\font plus
\BIBentryALTinterwordstretchfactor\fontdimen3\font minus \fontdimen4\font\relax}
\providecommand{\BIBforeignlanguage}[2]{{%
\expandafter\ifx\csname l@#1\endcsname\relax
\typeout{** WARNING: IEEEtran.bst: No hyphenation pattern has been}%
\typeout{** loaded for the language `#1'. Using the pattern for}%
\typeout{** the default language instead.}%
\else
\language=\csname l@#1\endcsname
\fi
#2}}
\providecommand{\BIBdecl}{\relax}
\BIBdecl

\bibitem{brown2020language}
T.~Brown, B.~Mann, N.~Ryder, M.~Subbiah, J.~D. Kaplan, P.~Dhariwal, A.~Neelakantan, P.~Shyam, G.~Sastry, A.~Askell \emph{et~al.}, ``Language models are few-shot learners,'' \emph{Advances in neural information processing systems}, vol.~33, pp. 1877--1901, 2020.

\bibitem{devlin2018bert}
J.~Devlin, M.-W. Chang, K.~Lee, and K.~Toutanova, ``Bert: Pre-training of deep bidirectional transformers for language understanding,'' \emph{arXiv preprint arXiv:1810.04805}, 2018.

\bibitem{foundationmodel}
R.~Bommasani, D.~A. Hudson, E.~Adeli, R.~Altman, S.~Arora, S.~von Arx, M.~S. Bernstein, J.~Bohg, A.~Bosselut, E.~Brunskill \emph{et~al.}, ``On the opportunities and risks of foundation models,'' \emph{arXiv preprint arXiv:2108.07258}, 2021.

\bibitem{xiao2023offsite}
G.~Xiao, J.~Lin, and S.~Han, ``Offsite-tuning: Transfer learning without full model,'' \emph{arXiv preprint arXiv:2302.04870}, 2023.

\bibitem{FedPET}
\BIBentryALTinterwordspacing
Z.~Zhang, Y.~Yang, Y.~Dai, Q.~Wang, Y.~Yu, L.~Qu, and Z.~Xu, ``{F}ed{PET}uning: When federated learning meets the parameter-efficient tuning methods of pre-trained language models,'' in \emph{Findings of the Association for Computational Linguistics: ACL 2023}.\hskip 1em plus 0.5em minus 0.4em\relax Toronto, Canada: Association for Computational Linguistics, Jul. 2023, pp. 9963--9977. [Online]. Available: \url{https://aclanthology.org/2023.findings-acl.632}
\BIBentrySTDinterwordspacing

\bibitem{radford2019language}
A.~Radford, J.~Wu, R.~Child, D.~Luan, D.~Amodei, I.~Sutskever \emph{et~al.}, ``Language models are unsupervised multitask learners,'' \emph{OpenAI blog}, vol.~1, no.~8, p.~9, 2019.

\bibitem{wei2021finetuned}
J.~Wei, M.~Bosma, V.~Y. Zhao, K.~Guu, A.~W. Yu, B.~Lester, N.~Du, A.~M. Dai, and Q.~V. Le, ``Finetuned language models are zero-shot learners,'' \emph{arXiv preprint arXiv:2109.01652}, 2021.

\bibitem{muennighoff2022crosslingual}
N.~Muennighoff, T.~Wang, L.~Sutawika, A.~Roberts, S.~Biderman, T.~L. Scao, M.~S. Bari, S.~Shen, Z.-X. Yong, H.~Schoelkopf \emph{et~al.}, ``Crosslingual generalization through multitask finetuning,'' \emph{arXiv preprint arXiv:2211.01786}, 2022.

\bibitem{houlsby2019parameter}
N.~Houlsby, A.~Giurgiu, S.~Jastrzebski, B.~Morrone, Q.~De~Laroussilhe, A.~Gesmundo, M.~Attariyan, and S.~Gelly, ``Parameter-efficient transfer learning for nlp,'' in \emph{International Conference on Machine Learning}.\hskip 1em plus 0.5em minus 0.4em\relax PMLR, 2019, pp. 2790--2799.

\bibitem{hu2021lora}
E.~J. Hu, Y.~Shen, P.~Wallis, Z.~Allen-Zhu, Y.~Li, S.~Wang, L.~Wang, and W.~Chen, ``Lora: Low-rank adaptation of large language models,'' \emph{arXiv preprint arXiv:2106.09685}, 2021.

\bibitem{qlora}
T.~Dettmers, A.~Pagnoni, A.~Holtzman, and L.~Zettlemoyer, ``Qlora: Efficient finetuning of quantized llms,'' \emph{arXiv preprint arXiv:2305.14314}, 2023.

\bibitem{rebuffi2017learning}
S.-A. Rebuffi, H.~Bilen, and A.~Vedaldi, ``Learning multiple visual domains with residual adapters,'' \emph{Advances in neural information processing systems}, vol.~30, 2017.

\bibitem{li2021prefix}
X.~L. Li and P.~Liang, ``Prefix-tuning: Optimizing continuous prompts for generation,'' \emph{arXiv preprint arXiv:2101.00190}, 2021.

\bibitem{qin2021learning}
G.~Qin and J.~Eisner, ``Learning how to ask: Querying lms with mixtures of soft prompts,'' \emph{arXiv preprint arXiv:2104.06599}, 2021.

\bibitem{lester2021power}
B.~Lester, R.~Al-Rfou, and N.~Constant, ``The power of scale for parameter-efficient prompt tuning,'' \emph{arXiv preprint arXiv:2104.08691}, 2021.

\bibitem{liu2021p}
X.~Liu, K.~Ji, Y.~Fu, W.~L. Tam, Z.~Du, Z.~Yang, and J.~Tang, ``P-tuning v2: Prompt tuning can be comparable to fine-tuning universally across scales and tasks,'' \emph{arXiv preprint arXiv:2110.07602}, 2021.

\bibitem{an2022input}
S.~An, Y.~Li, Z.~Lin, Q.~Liu, B.~Chen, Q.~Fu, W.~Chen, N.~Zheng, and J.-G. Lou, ``Input-tuning: Adapting unfamiliar inputs to frozen pretrained models,'' \emph{arXiv preprint arXiv:2203.03131}, 2022.

\bibitem{ding2023dc}
Y.~Ding, C.~Niu, F.~Wu, S.~Tang, C.~Lyu, and G.~Chen, ``Dc-ccl: Device-cloud collaborative controlled learning for large vision models,'' \emph{arXiv preprint arXiv:2303.10361}, 2023.

\bibitem{dong2023lambo}
L.~Dong, F.~Jiang, Y.~Peng, K.~Wang, K.~Yang, C.~Pan, and R.~Schober, ``Lambo: Large language model empowered edge intelligence,'' \emph{arXiv preprint arXiv:2308.15078}, 2023.

\bibitem{shen2023large}
Y.~Shen, J.~Shao, X.~Zhang, Z.~Lin, H.~Pan, D.~Li, J.~Zhang, and K.~B. Letaief, ``Large language models empowered autonomous edge ai for connected intelligence,'' \emph{arXiv preprint arXiv:2307.02779}, 2023.

\bibitem{modelcompression}
C.~Xu and J.~McAuley, ``A survey on model compression and acceleration for pretrained language models,'' in \emph{Proceedings of the AAAI Conference on Artificial Intelligence}, vol.~37, no.~9, 2023, pp. 10\,566--10\,575.

\bibitem{FDMA}
H.~Mathur and T.~Deepa, ``A survey on advanced multiple access techniques for 5g and beyond wireless communications,'' \emph{Wireless Personal Communications}, vol. 118, pp. 1775--1792, 2021.

\bibitem{wirelessDRL}
N.~C. Luong, D.~T. Hoang, S.~Gong, D.~Niyato, P.~Wang, Y.-C. Liang, and D.~I. Kim, ``Applications of deep reinforcement learning in communications and networking: A survey,'' \emph{IEEE Communications Surveys \& Tutorials}, 2019.

\bibitem{PPO}
J.~Schulman, F.~Wolski, P.~Dhariwal, A.~Radford, and O.~Klimov, ``Proximal policy optimization algorithms,'' \emph{arXiv preprint arXiv:1707.06347}, 2017.

\bibitem{GAE}
J.~Schulman, P.~Moritz, S.~Levine, M.~Jordan, and P.~Abbeel, ``High-dimensional continuous control using generalized advantage estimation,'' \emph{arXiv preprint arXiv:1506.02438}, 2015.

\bibitem{RLintro}
R.~S. Sutton and A.~G. Barto, \emph{Reinforcement Learning: An Introduction}.\hskip 1em plus 0.5em minus 0.4em\relax MIT press, 2018.

\bibitem{liyueheng}
Y.~Li, G.~Xie, and Z.~Lu, ``Difference advantage estimation for multi-agent policy gradients,'' in \emph{International Conference on Machine Learning}.\hskip 1em plus 0.5em minus 0.4em\relax PMLR, 2022, pp. 13\,066--13\,085.

\bibitem{CTDE}
P.~K. Sharma, R.~Fernandez, E.~Zaroukian, M.~Dorothy, A.~Basak, and D.~E. Asher, ``Survey of recent multi-agent reinforcement learning algorithms utilizing centralized training,'' in \emph{Artificial Intelligence and Machine Learning for Multi-Domain Operations Applications III}, vol. 11746, 2021, pp. 665--676.

\bibitem{size}
T.~Brown, B.~Mann, N.~Ryder, M.~Subbiah, J.~D. Kaplan, P.~Dhariwal, A.~Neelakantan, P.~Shyam, G.~Sastry, A.~Askell \emph{et~al.}, ``Language models are few-shot learners,'' \emph{Advances in neural information processing systems}, vol.~33, pp. 1877--1901, 2020.

\bibitem{FDD}
K.~I. Pedersen, G.~Berardinelli, F.~Frederiksen, P.~Mogensen, and A.~Szufarska, ``A flexible 5g frame structure design for frequency-division duplex cases,'' \emph{IEEE Communications Magazine}, 2016.

\bibitem{IPPO}
C.~Yu, A.~Velu, E.~Vinitsky, Y.~Wang, A.~Bayen, and Y.~Wu, ``The surprising effectiveness of {PPO} in cooperative, multi-agent games,'' \emph{arXiv preprint arXiv:2103.01955}, 2021.

\end{thebibliography}

\end{document}